\documentclass[fleqn,10pt]{wlscirep}
\usepackage[utf8]{inputenc}
\usepackage[T1]{fontenc}
\usepackage{lineno}

\usepackage{array}

\usepackage{comment}
\usepackage{textcomp}
\usepackage{gensymb}
\usepackage{url}

\title{Full Field Digital Mammography Dataset from a Population Screening Program}

\author[1,*]{Edward Kendall}
\author[2]{Parham Hajishafiezahramini}
\author[2]{Matthew Hamilton}
\author[3]{Gregory Doyle}
\author[1]{Nancy Wadden}
\author[2]{Oscar Meruvia-Pastor}

\affil[1]{Faculty of Medicine, Memorial University of Newfoundland, St John's, NL, Canada}
\affil[2]{Department of Computer Science, Faculty of Science, Memorial University of Newfoundland, St John's, NL, Canada}
\affil[3]{Cancer Care, Newfoundland and Labrador Health Services, St John's, NL, Canada}

\affil[*]{Corresponding author: Matthew Hamilton (mhamilton@mun.ca)}

\begin{abstract}

Breast cancer presents the second largest cancer risk in the world to women. Early detection of cancer has been shown to be effective in reducing mortality. Population screening programs schedule regular mammography imaging for participants, promoting early detection. Currently, such screening programs require manual reading. False-positive errors in the reading process unnecessarily leads to costly follow-up and patient anxiety.
Automated methods promise to provide more efficient, consistent and effective reading. To facilitate their development, a number of datasets have been created. Such datasets can aid in learning-based development but many are not publicly available and do not draw directly from population screening programs. With the aim of specifically targeting population screening programs, we introduce NL-Breast-Screening, a dataset from a Canadian provincial screening program. The dataset consists of 5997 mammography exams, each of which has four standard views and is biopsy-confirmed. Cases where radiologists' reading was a false-positive are identified. NL-Breast is made publicly available as a new resource to promote advances in automation for population screening programs.

\end{abstract}
\begin{document}

\flushbottom
\maketitle

\thispagestyle{empty}


\section*{Introduction}

Breast cancer prevalence increases with age and health authorities recommend that women over 40 (or 50 in some jurisdictions) receive an X-ray mammogram every two years. As a result, in Canada and the USA, about 40 million exams are performed each year. Of these, two million exams (5\%) are deemed suspicious by the radiologist and referred for additional procedures. However, only 270,000 cases of breast cancer are reported each year. Assuming that all these cases are represented in the mammograms, the disease incidence in the mammograms is no greater than 0.7\% (270k/40M).  Ironically, although it appears that there is a marked tendency to over-diagnose, screening programs still miss up to 20\% of the resident disease \cite{Lehman2017NationalConsortium}. Attempts to mitigate errors include double reading and artificial intelligence (AI) assistance. These efforts improved accuracy but increased cost; they did not succeed in eliminating the false negative and false positive cases. In response to this diagnostic challenge, many research groups applied advanced automated detection and classification schemes to the problem of breast cancer detection in mammography\cite{Guo2023ArtificialCancer}. The results, while promising, have not eclipsed those of existing programs \cite{Michalopoulou2023, Hollingsworth2019CentreScreening}.  

Since the onset of screening programs in the 1980’s there have been many technological developments. The first was optimization of x-ray beam quality through the use of appropriate anode material and aperture filters. During this period film technology received a great deal of attention, very fine structured films provided superior spatial resolution at the cost of a small dose increase. Film was supplanted by indirect technology, which in turn was supplanted by digital technology. Digital technology brought an opportunity to standardize data formats. Mammograms are now stored using the medical-standard DICOM image format \cite{Bidgood1997UnderstandingImaging}. This format is information rich in that it includes details of equipment settings and of automated data processing routines. 
Detector development continues to deliver ever higher resolution and ever more sensitive detectors. In addition, line source x-ray generators promise to reduce scatter and lower dose. To assist with the challenge of overlying tissue, panoramic cameras are now used to help with depth perception and increase conspicuity. There are now thousands of terabytes of mammography data, representing just about every variation of normal and pathological breast anatomy. This has proven to be an attractive resource for software engineers, who have produced a steady stream of “computer aided diagnostic” (CAD) applications to assist in identifying markers of disease. 

Early offerings used intensity measures to identify bright clusters in mammograms \cite{doi:10.1056/NEJMoa066099}. When present, these represent calcifications that, in certain configurations, are associated with carcinoma development. However, not all cancer containing breasts feature calcifications \cite{Naseem2015MammographicAnalysis, Azam2021MammographicCancer}. So, the next focus was on identifying masses. These were detectable as discrete, anomalous areas of intermediate brightness. They represent a large class of breast cancer known as ductal carcinoma in situ \cite{Menon2024BreastCancer}. Other cancers have a more complex structure, often with tendrils or diffuse edges that extend into seemingly normal tissue \cite{Menon2024BreastCancer}. These were more difficult to detect using software algorithms, but were an important entity since they represent a second class of breast tumour known as invasive ductal carcinoma\cite{doi:10.7326/0003-4819-138-3-200302040-00008}. There are even more challenging cases. When the native breast tissue is fibrous, the contrast between normal and cancer tissues, already very subtle, is reduced \cite{doi:10.7326/0003-4819-138-3-200302040-00008}. Younger women tend to have these type of breasts; the detection efficiency is reduced for this age group meaning that some programs do not include women under fifty years of age \cite{doi:10.7326/0003-4819-138-3-200302040-00008}. 

As the algorithms have grown in complexity it has been more difficult to realize diagnostic gains in breast cancer screening programs. In addition, it is important not to increase the resource demands on this already expensive program \cite{CanadianTaskForceonPreventiveHealthCareBreastHttps://canadiantaskforce.ca/breast-cancer-update-draft-recommendations/}. Early studies determined that the use of CAD reduced the overall efficiency of a screening program as the radiologist was faced with many false positive findings that had to be cleared \cite{Dromain2013Computed-aidedCancer, PMID:23104393, Houssami2009EarlyArticle, Houssami2009BreastMammography}. Subsequent studies have provided a mixed bag of results. The performance of expert screeners did not improve with the use of CAD, while that of less experienced mammographers did. Despite this, software developers have pressed on, focusing their attention on differentiating benign and cancerous findings and on grading the severity of tumours found in an image. This focus has produced very complex CAD programs whose use requires specific training by the radiologist. Even though the number of false positives increased with CAD, its use in screening mammography is now widespread. 

CAD development has slowed somewhat and that may be attributed to the complexity of the remaining tasks. Generally in software development early applications develop quickly but have a narrow range of utility. As the scope of the undertaking increases, so too does the number of potential failure points. In very complex situations, such as differentiating tumour grade, there is an exponential increase in the number of potential failure points and a corresponding increase in the time required to resolve them. A second barrier to development is the requirement that software which is used in diagnosis must be approved by the cognate national authority. This authorization requires extensive independent testing, and that has highlighted another problem: the shortage of high quality testing datasets. 

\section*{Review of Existing Datasets}

Previously, it was noted that a great deal of development was done using two legacy mammographic datasets, the MIAS and DDSM \cite{Matheus2011OnlineSchemes}. There are now several others (reviewed below) but only two which contain a normal representative distribution of pathology reflecting the population of a screening program \cite{Halling-Brown2021OPTIMAMData, Dembrower2020}; these datasets are nevertheless not readily available for public download. The popular DDSM dataset has an abnormal distribution in that there are 695 normal findings and 1784 abnormal cases. This requires a sub-sampling strategy to achieve a realistic distribution, and even with this there are not enough normal cases. Of course, there are correlation strategies to compensate for skewed datasets, but there is always the danger that an algorithm will be over-specified for a particular set of data and will perform very differently when the distribution changes. 

A second problem with all the available datasets is that they represent a range of acquisition technologies. For example, the DDSM and MIAS sets are digitized film images. Others contain a mixture of scanned, indirect and direct digital images. The file types include JPEG, TIFF, PGM and BMP. Strangely, few of them provide the widely used DICOM format. This provides additional problems when trying to license software for medical applications as the test image format becomes a variable for consideration. Aside from the format itself, variations in spatial resolution and bit depth present challenges to programmers. 

On the spatial resolution front, software that is designed to recognize clumps of small calcifications will overlook the fact that those calcifications may be present in a 0.5 mm pixel. On the other hand, discontinuities in tracts at 0.01 mm resolution may be misinterpreted as an anomaly. 

Bit depth is useful to distinguish subtle differences in intensity. However, subtle differences are present in every breast image so the expanded grey scale range may not be particularly useful. The combination of high resolution and large bit depth creates a very large file that may increase processing times substantially. Reducing the bit depth to 8 bit from 16 bit lessened the processing time from hours to minutes \cite{Jiang2023Space-to-speedProcessing}.

The file formats, resolution and size present real but not insurmountable barriers for efficient software development. However, many existing public datasets do not contain some of the important instrument information such as anode type, operating energy, exposure time, window width, and filters and processing. This data is very useful in arriving at a consistent protocol for post-acquisition processing screening images. The DICOM file format contains this information as a header in each image file, negating the need to cross-reference a CSV file. For example, using information in the DICOM header, a classification detection algorithm can be adjusted for instrument variables including resolution and bit depth. Applications that can use the DICOM information can be deployed on Health Region networks where multiple camera types may be present. On the other hand, the DICOM dataset is large and may require parallel computing resources to process efficiently.


One of the main challenges with deep learning approaches to detection is the requirement for a sufficiently large training dataset in order to achieve high detection rates that could be ultimately generalized to a broad population of patients. More patient image data, captured from a variety of imaging conditions can likely help with this generalization. One issue with the datasets for breast screening is there are not so many datasets specified on breast cancer screening containing normal and suspicious(probably cancerous) cases. The datasets that have been frequently used in different studies are:

Digital Database for Screening Mammography (DDSM) contains 10,480 images (2,620 exams)\cite{Heath2001}, though the dataset is a digitalized scan of mammography. Suspicious cases are verified are benign or malignant. Masses and calcifications are identified and localized in malignant images and further categorized according to type. Further, breasts are categorized in terms of density using ACR BI-RADs system. The portion of abnormal cases in this dataset is much higher than what would normally be found in a screening population, suggesting that this dataset does not directly reflect a screening program. Nevertheless, it could be used for screening programs since the images are divided into normal and suspicious cases.


CBIS-DDSM  (Curated Breast Imaging Subset of DDSM) dataset\cite{Lee2017b} consists of native size, enhanced mass segmentation, and pathologic diagnosis integrated into the training data, structured akin to contemporary computer vision datasets. It contains 753 instances of calcification and 891 instances of masses. This dataset cannot be used for the screening program classification as all the cases are cancerous. However, it can be used to discriminate between benign tumours and invasive ones. 

MIAS: The MIAS dataset was first published in 1994 and has been widely used in research \cite{sucklingetal._2015}. The MIAS dataset contains digitized versions of film images scanned at 50 micron resolution. The files are curated with information on the type of anomalies found in each image, as well as a composition rating of the background tissue. The MLO-view image files are PGM formatted, their overall size depends on the size of the breast tissue scanned. The dataset contains 161 pairs of images.

The InBreast dataset \cite{Moreira2012} was introduced in 2011. It contains 115 cases, totalling 410 images. Ninety cases were women with both breasts affected, each with four images, while 25 cases involved mastectomy patients; they only had two images per patient. Specialists in this dataset provide accurate contours (ROI) in XML format.

VinDr \cite{Nguyen2023} was gathered in Vietnam from 5,000 mammography exams; each patient sample contains two medial lateral oblique  and two cranial caudal views, resulting in 20,000 images in total. It used  double-reads and resolved disagreements through arbitration. This dataset includes the Breast Imaging Reporting and Data System (BI-RADS) score, reported breast density and also annotates non-benign findings with their category, location, and BI-RADS assessment. The positive cases do not appear to be biopsy confirmed however, thus may contain some false positives.

CMMD : The Chinese Mammography Database (CMMD) \cite{Cai2023} contains 3712 mammographic images from 1775 patients, divided to 2 separate parts. CMMD1 consists of 1026 cases (2214 images) with biopsy-confirmed benign or malignant tumours, while CMMD2 includes 1498 mammographs for 749 patients with known molecular subtypes. This dataset was constructed to enhance the diversity of mammography data and foster advancements in related fields. Each patient mammography includes craniocaudal (CC) and mediolateral oblique (MLO) projection images, stored as 8-bit grayscale DICOM files at 2294 × 1914 pixels resolution. Despite its value, the dataset's limitations include its relatively small sample size and lack of marked regions of interest (ROI).

The RSNA dataset \cite{rsna-breast-cancer-detection} contained 54,713 digital mammograms from almost 8,000 patients. Of these, 570 were right breast positive, 588 were left breast positive, and 2 were positive in both breasts. This dataset is quite extensive but has only 2.1\% percent positive cases, making it more similar to a screening population dataset than many others. As we reviewed random samples from the dataset, we observed instances where the indicated laterality of the breast was incorrect. For example, some images labeled as left were, in fact, right breast. This dataset is publicly available and was the subject of a recent Kaggle data challenge.

OPTIMAM (OMI-DB) \cite{Halling-Brown2021OPTIMAMData} is a very large FFDM-based dataset containing data from 172,282 women screened in the UK. Interval cancers are tracked as well as positive cases biopsy-confirmed. The dataset also contains pixel level annotations of tumour features. The dataset is available by request on a per-project basis, subject to committee approval.

The Swedish screening dataset (CSAW) \cite{Dembrower2020} also provides a sizable FFDM dataset directly drawn from a screening program in Sweden, containing data for 499.807 women. The dataset contains sigificant metadata, including biospy-verification and histological and tumour information. This data also only provides access by request and cannot be readily downloaded and used.  

EMory BrEast Imaging Dataset (EMBED) \cite{Jeong2023TheImages} also presents another large scale FFDM dataset, containing 116,000 patients, from racially diverse backgrounds with a large (42$\%$) African American representation, in contrast to other datasets which are more homogenous. This dataset does not appear limited strictly to screening populations, thus not reflective of these programs directly statistically. A portion (20 percent) of this dataset is available for download, other access must be requested.

ADMANI: Annotated Digital Mammograms and Associated Non-Image Datasets \cite{doi:10.1148/ryai.220072} is another significant dataset collected in Australia containing 629,863 patients from 1 million screening episodes. Some portion of this data also appears in the RSNA Kaggle challenge dataset mentioned above \cite{rsna-breast-cancer-detection}. Outside of this portion, however the dataset does not appear to be available.

 The Breast Cancer Digital Repository (BCDR) is a comprehensive database aimed at advancing breast cancer detection and diagnosis methods. It contains cases from 1734 patients, including mammography and ultrasound images, clinical history, lesion segmentation, and image-based descriptors. The repository is subdivided into Film Mammography-based (BCDR-FM) and Full Field Digital Mammography-based (BCDR-DM) branches. BCDR-FM includes cases from 1010 patients, while BCDR-DM is still under construction and includes cases from 724 patients. Both repositories provide annotated patient cases, lesion outlines, and clinical data, making them valuable resources for research and training in breast cancer imaging.

\begin{figure}[ht]
\centerline{\includegraphics[width=0.97\linewidth]{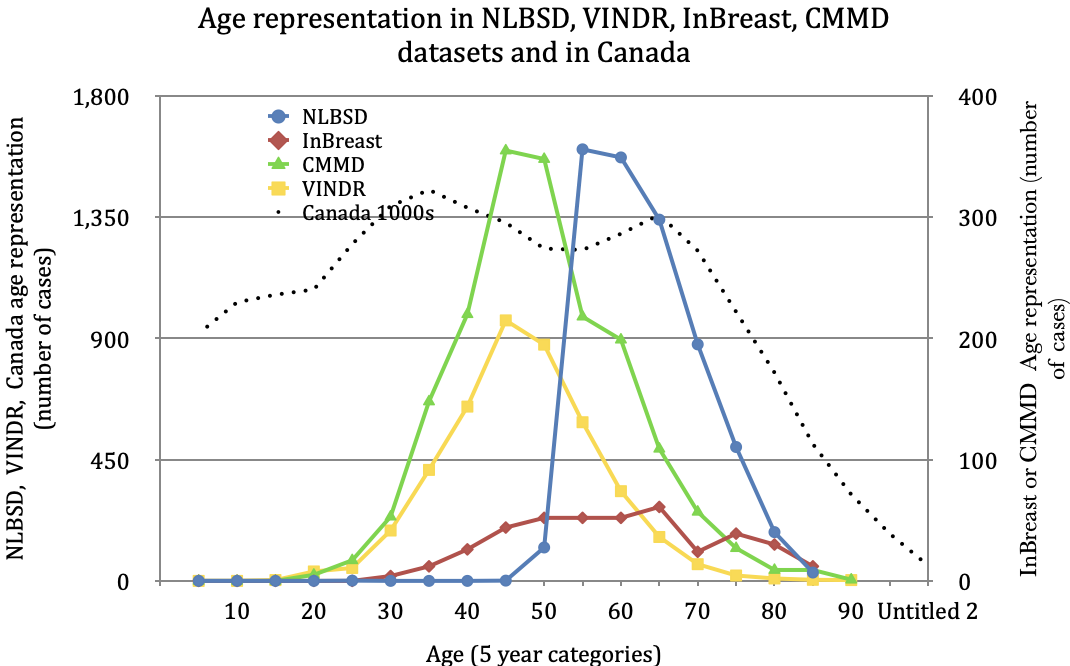}}
\caption{Age representation in NLBS, VINDR, InBreast, CMMD Datasets, and in Canada (women, 2024 \cite{StatisticsCanada2024TableGender}).}
\label{figure:AgeRepresentation}
\end{figure}

Like many software developers we have laboured with poorly constructed, out of date datasets. In addition, the prior-probability situation was constantly raised in publication review. If there are 10 cancer presentation variations and if, in total, the incidence is 0.5\% this means that the data set must contain at least fifty thousand normal cases. Of course the actual number will have to be higher to ensure that the variations on cancer type and the variations on normal are fully represented. 

Here we report our efforts to construct a mammography dataset with a realistic distribution in a medically relevant format. 

\section*{Methods}
Breast screening images were obtained through the auspices of the Newfoundland and Labrador Health Services (NLHS) breast screening program. Patient consent was obtained by the screening program. The data was collected on GE digital Senograph Essential flat field digital devices (manufactured 2008-10). All cases listed as positive were confirmed through additional diagnostic procedures and by laboratory analyses. False positive cases were those initially identified as suspicious by the screening radiologist, but later determined to be normal by additional diagnostic procedures and laboratory analysis. Normal cases are those deemed normal by the screening radiologist and subsequently determined to be normal by lack of interval cancer. These designations were provided by the screening program and adopted here so that computer based classification schemes might easily translate to the clinic. In all, 26,988 images representing 5997 cases were obtained. Of these, 5935 cases were determined to be unique. The balance were repeat exams as determined by the anonymizing identifier.

\begin{figure}[ht]
\centerline{\includegraphics[width=1.07\linewidth]{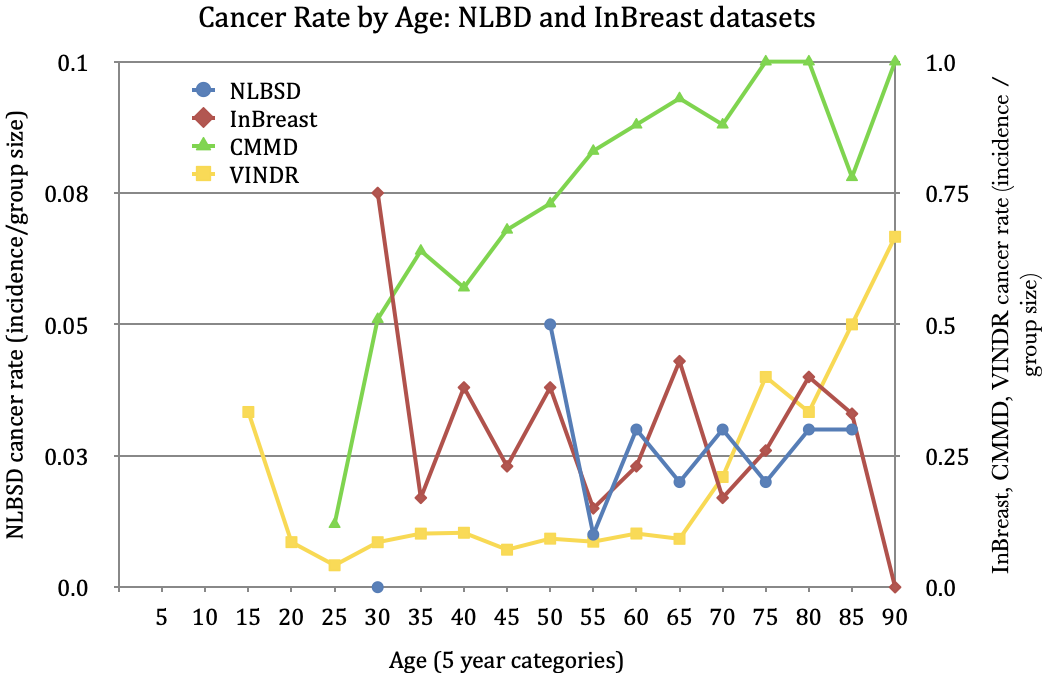}}
\caption{Cancer rates by age in various mammography datasets.}
\label{figure:CancerByAge}
\end{figure}

\begin{table}
\centering

\begin{tabular}{|l|cc|c|>{\centering\arraybackslash}p{0.10\linewidth} |}

\hline
&  All&Normal& False positive& Positive\\
\hline
\hline
 &  && & \\
 Cases&  5997&4332& 1516&149\\
 Images&  26988&19942& 6394&652\\
 Images/case&  &4.6& 4.2&4.4\\
 Mean age (y)&  &61.9 $\pm$ 7.3& 58.9 $\pm$ 7.0&60.2 $\pm$ 7.1\\
 Median age (y)&  &60.1 & 58.0 & 61.0 \\
 75\%ile age (y)&  &65.7 & 63.0 & 67.0 \\
 Exams/ case&  &6.12 & 4.12 & 5.13 \\
 Right breast images&  13549&10017& 3208&324\\
 Left breast images&  13439&9925& 3186&328\\
 Cranial caudal view&  13800&9656& 3206&326\\
 Medial lateral oblique view&  13188&10286& 3188&326\\
 4.39Mpx& 8267& 6291& 1848&128\\
 7.34Mpx& 18695& 13636& 4536&523\\
\hline
\end{tabular}
\caption{Description of the NLBS dataset}
\label{table:NLBSdescript}
\end{table}


The  NL-Breast-Screening (NLBS) dataset contains native-resolution anonymized DICOM (16 bit, 2394 x 3062 [7.34 megapixels] and 1915 x 2295[4.39 megapixels]) files. The images were obtained using similar equipment.  Mean age in the 4.39 Mpixel set versus the 7.34 Mpixel set was significantly lower ($p < 0.001$) in the false positive (58.4 $\pm$ 6.9 vs. 59.1 $\pm$ 7.1)  and normal groups (59.4 $\pm$ 7.1 vs. 60.6 $\pm$ 7.1) but not in the positive group (62.2 $\pm$ 7.6 vs. 61.8 $\pm$ 7.2).



This dataset is large enough that meaningful classification studies can be performed using realistic prior probabilities. The incidence of cancer by age in this dataset is more reflecitve of that in a screeing population compared with other readily-available datasets (Figure \ref{figure:CancerByAge}). The classification exercise then must struggle with the bias to normalcy that is similar to the situation in the clinic. There are 19942 normal images representing 4332 cases (4301 unique), 1516 false positive cases (1484 unique) containing 6394 images, and 149 unique cancer cases with 652 images (See Table \ref{table:NLBSdescript} for a more complete summary of the dataset). 

The vast majority of these images were obtained using the Rhodium anode (25655) with a mean energy of  29 $\pm$ 0.7 keV, images obtained using the Molybdenum anode (2312) used a mean energy of 27 $\pm$ 0.6 keV. In Table 1, it may be noted that the mean age of the "positive" group was 61.9 $\pm$ 7.3, that of the false positive group 58.9 $\pm 7.0$ and that of the normal group 60.2 $\pm$ 7.1. The balance of image laterality representation (R/L) was similar among the groups: abnormal 324/328, false positive 3208/3186 and normal 10017/9925. Similarly, the view MLO/CC (13800/ 13188) distribution was similar for each group: positive - 326/326, false positive 3206/3188, and normal 9656/10286. 

It may be argued that the number of confirmed positive cases is insufficient to represent all potential cancer presentations. Our plan is to continue collecting data to address this potential deficiency. In the meantime, one can also augment the dataset with images from other available dataset (e.g. CMMD, VinDr or RSNA datasets). Having said that, it is important to note that the populations represented in the Chinese and Vietnamese datasets may have biophysical characteristics that differ from, the Canadian population represented in our dataset. For example, our NLBS participants are somewhat older than those in VinDr and CMMD datasets (See Figure \ref{figure:CancerByAge}). This reflects the screening guidelines in Canada ( $\ge$ 50   $\leq$ 70 and the age distribution of the Canadian female population. 

Annontations which classify the breast as dense, fatty or intermediate and  positive-case features classified as masses, calcifications or architectural distortions are not included. However, the dataset is well verified since all normal cases are verified as cancer free for at least two years, the false positive group was verified by biopsy or additional diagnostic imaging and the positive group was biopsy confirmed.

The NLBS dataset is available for research purposes. 
The DICOM dataset is approximately 350 GB, representing a broad range of pathology and normalcy. The great advantage of including the DICOM headers is that these can be interrogated to identify sequential scans, age, and a host of equipment factors that can inform a classification investigation. 

\section*{Data Records}

For the dataset presented, DICOM images and radiologists findings of the dataset have been made available at the Federated Research Data Repository (FRDR) \cite{KendallNewfoundlandRepository}. In parallel, a submission for image hosting at The Cancer Imaging Archive (TCIA) has also been made.

At the case level, the set of exam images are classified as either positive, normal or false positive. At the highest level, the dataset is organized therefore into three main folders, \texttt{positive/}, \texttt{normal/} or \texttt{false positive/}. Beneath each of these top-level folders is a set of folders, each representing a patient case with the finding of the top level folder named with a de-identified label. Within this folder are all the images for this particular case, contained within two more sub folders, \texttt{CC/} and \texttt{MLO/}, containing two images each for the craniocauda (CC) and mediolateral oblique (MLO) views, respectively.  

Patient-case level findings are stored in a CSV file \texttt{NLBSP-meta.csv}. There are 6 columns in the CSV file:

\begin{itemize}
 \setlength\itemsep{0em}
\item \texttt{File Path}: Relative path of DICOM file, based on the file path structure described above. 
\item \texttt{Image Laterality}: Laterality of the DICOM image (also reflected in the file path).
\item \texttt{View Position}: CC or MLO (also reflected in the file path).
\item \texttt{Age}: Age of the screening participant in the image at location specified in \texttt{File Path}.
\item \texttt{Cancer}: This column is \texttt{1} if cancer was found, and \texttt{0} for a normal case.
\item \texttt{False Positive}: This column will read \texttt{1} for an image that was positive, but later confirmed as negative. Otherwise the column will be \texttt{0}.
\end{itemize}

Within the DICOM file specified at \texttt{File Path}, a variety of other informative fields are populated, however any identifying fields have been blanked for privacy protection purposes.

\section*{Technical Validation}

The data was anonymized to ensure that individually identifiable health information or PHI was completely removed. In addition, image content was reviewed manually to ensure no patient data remained in mammography images or other images files.

\section*{Summary}
Herein is a description of a novel dataset for mammogram-based AI studies. The FFDM data was collected on similar equipment consequently resolution and image presentation are consistent throughout. The dataset is fully validated as to outcome. However, it lacks image annotation and textual summative findings. The age demographic is well understood and the ethnicity is assumed to be European. The dataset is provided for research purposes and is available at doi:. 

\bibliography{MDreferences,DataPaperReferences}  

\end{document}